# Deep-Learning-Based Markerless Pose Estimation Systems in Gait Analysis: DeepLabCut Custom Training and the Refinement Function

**Giulia Panconi[1], Stefano Grasso[2], Sara Guarducci[3], Lorenzo Mucchi[3], Diego Minciacchi[1], Riccardo Bravi[1]**

[1]Department of Experimental and Clinical Medicine, University of Florence, Italy
[2]Department of Physiology and Pharmacology, SAPIENZA University of Rome, Italy
[3]Department of Information Engineering, University of Florence, Italy

## Abstract

The current gold standard for the study of human movement is the marker-based motion capture system that offers high precision but constrained by costs and controlled environments. Markerless pose estimation systems emerge as ecological alternatives, allowing unobtrusive data acquisition in natural settings. This study compares the performance of two popular markerless systems, OpenPose (OP) and DeepLabCut (DLC), in assessing locomotion.

Forty healthy subjects walked along a 5 meters walkway equipped with four force platforms and a camera. Gait parameters were obtained using OP "BODY_25" Pre-Trained model (OPPT), DLC "Model Zoo full_human" Pre-Trained model (DLCPT) and DLC Custom-Trained model (DLCCT), then compared with those acquired from the force platforms as reference system. Our results demonstrated that DLCCT outperformed DLCPT and OPPT, highlighting the importance of leveraging DeepLabCut transfer learning to enhance the pose estimation performance with a custom-trained neural networks. Moreover, DLCCT, with the implementation of the DLC refinement function, offers the most promising markerless pose estimation solution for evaluating locomotion. Therefore, our data provide insights into the DLC training and refinement processes required to achieve optimal performance.

This study offers perspectives for clinicians and practitioners seeking accurate low-cost methods for movement assessment beyond laboratory settings.

# Introduction

The development of accurate and thorough analysis is an essential aspect for the study of human motion. The actual gold standard for measuring human biomechanics is marker-based motion capture. This approach offers precise tracking accuracy down to sub-millimeter levels (Kessler et al., 2019, Miranda et al., 2013, Wang et al., 2021), making it particularly well-suited for the study of motor control (Cook et al., 2007, Metcalf et al., 2008, Bravi et al., 2022, Hartmann et al., 2023). In sport science, marker-based approach is widely employed to meticulously examine complex movements during athletic performance (Pfile et al., 2013, Weeks et al., 2015, van der Kruk et al., 2018). Likewise, in clinical settings, the implementation of marker-based motion capture analysis addresses the constrains of clinicians' subjective qualitative assessments, which often suffer from a lack of inter-rater reliability (Li et al., 2018, Moro et al., 2020, Bravi et al., 2021, Stenum et al., 2021b).

Nevertheless, marker-based tools are characterized by expensive cost, requirement of specially trained operators, and controlled environment obligation, which constrain their applicability. Furthermore, using physical markers placed on the skin can introduce errors due to their mobility during movement (i.e., soft tissue artifacts), potentially compromising data confidence (Peters et al., 2010, Camomilla et al., 2017).

Altogether, marker-based characteristics may contribute to a reduction in the richness and confidence of data gathering, as well as an alteration in movement patterns (Gorton et al., 2009, Johnson et al., 2018, Fleisig et al., 2021, Lahkar et al., 2022). These limitations underscore the urgent need of more eco-friendly alternatives for human movement assessment (Wade et al., 2022, Moro et al., 2022).

Markerless pose estimation represents a challenging computer vision task aimed at determining the location, orientation, and motion of a subject without physical markers. Recent developments in neural networks have led to the creation of more efficient and robust pose estimation methods, largely addressing the limitations of the marker-based approach. Markerless pose estimation procedure entails the identification of keypoints corresponding to specific body parts to reconstruct the subject through a simple video recording, thus representing an unobtrusive procedure for capturing extensive data in ecologically natural conditions (Mathis et al., 2020, Stenum et al., 2021b, Wade et al., 2022, Scott et al., 2022). So far, several open-source software programs have been implemented (Abadi et al., 2016, Cao et al., 2018, Mathis et al., 2018, Bazarevsky et al., 2020, Fang et al., 2023). To provide an in-depth examination of some of the most used markerless software programs, we decided to focus on two popular tools: OpenPose (OP) and DeepLabCut (DLC), on which a few previous comparative attempts have already been carried out (Needham et al., 2021, Washabaugh et al., 2022, Van Hooren et al., 2022). On one hand, OP is one of the most widely used tool for markerless motion capture, owing to its off-the-shelf pre-trained neural network (Cao et al., 2018, Slembrouck et al., 2020, Yamamoto et al., 2022, Hii et al., 2023). These pre-trained neural networks are algorithms that commonly employ models trained on extensive datasets containing millions of human images to accurately track specific keypoints. On the other hand, the DLC software provides an artificial neural network that, starting from a pre-trained model, can be trained to recognize the locations of manually designated landmarks, and develop a custom-trained model based on a specific dataset (Mathis et al., 2018, Nath et al., 2019,

Kosourikhina et al., 2022). The differences between pre- and custom- trained models lie in their training processes. Pre-trained models are initially trained on large datasets for general tasks and are subsequently fine-tuned for specific domains. In contrast, custom-trained models are either built from scratch or adapted to learn task-specific patterns more effectively.

Despite the wide potentialities and the expanding fields of application of markerless pose estimation systems, only a few comparative accuracy studies have used this approach to assess gait performance (Needham et al., 2021, Washabaugh et al., 2022), a typical movement behavior analyzed in clinical settings (Fritz et al., 2015). In addition, these studies used DLC's pre-trained model without implementing a custom-trained model, which would enable a more effective utilization of the advantageous transfer learning properties of this software. Transfer learning enables the reutilization and refinement of a model trained on one task for another correlated task, allowing high-performance landmark tracking with fewer images. Moreover, Needham et al. (2021) and Washabaugh et al. (2022) employed data from multiple cameras rather than utilizing the typical single 2D camera which is more commonly available in non-laboratory setups (Verlekar et al., 2019, Li et al., 2019, Van Hooren et al., 2022).

Therefore, our goal is to conduct a comparative assessment of the accuracy of OP and both pre- and custom-trained models of DLC in measuring healthy subjects' locomotion. We then compared the temporal gait metrics collected from markerless systems with those obtained from the reference system represented by force platforms. Furthermore, we also aim to analyze and evaluate the DLC training and refinement processes, which encompass outlier extraction and manual annotation correction, required to achieve optimal performance.

Our study will contribute to a progress in comprehension of open-source markerless pose estimation tools, potentially extending their applications beyond laboratory environments, thereby also benefiting clinicians and practitioners.

## Methods

*Experimental setup and design*

Forty healthy subjects (21 males, age: 23.95 ± 5.42 years, height: 1.80 ± 0.07 m, weight: 74.05 ± 10.34 kg; 19 females, age: 22.32 ± 1.63 years, height: 1.67 ± 0.05 m, weight: 58.37 ± 5.82 kg) were recruited from the University of Florence, as they volunteered to participate and provided written informed consent. They wore their own clothing and footwear and did not present any gait alteration.

Subjects walked along a 5-meter indoor walkway equipped with four force platforms (INFINI-T, single platform sensitive area: 60 × 40 cm, capacity for each module ± 8000 N, sensitivity/resolution: 16 bits over the selected range, sampling frequency: 1000 Hz, BTS Bioengineering Corp., Italy) installed in the center of the walkway to record middle gait steps (Fig. 1A). The sagittal plane views of the walking sequence were recorded by a RGB camera (Vixta, sampling frequency: 25 fps, resolution: 640 × 480, BTS Bioengineering Corp., Italy), which was time-synchronized with the force platforms, as both systems were recorded using BTS Smart Clinic software. The camera was mounted on a tripod with a height of 1 m and located 1.4 m from the center of the walkway (Fig. 1A).

Subjects were instructed to start walking upon receiving the signal and to proceed at their own pace along the indoor walkway. They were asked to complete five laps, navigating back and forth in front of the camera for a total of ten times (Fig. 1A). This allowed for the recording of both the right and left sides of the tested subject.

*Force Platform Signal Processing*

Raw force signals obtained from force platforms were processed by applying a second-order Butterworth low-pass filter with a 20 Hz cutoff to reduce noise (May & Willwacher, 2019, Horsak et al., 2020). To obtain gait temporal parameters, the heel-contact (HC) event, when the heel first contacts the walking surface, and the toe-off (TO) event, when the foot is lifted from the platform surface, were determined. The HC event was identified as the first sample of the vertical ground reaction force above the zero line (Hansen et al., 2002). The TO event was identified as the first sample when the vertical ground reaction force returns to zero.

*Video Files Processing*

Video files obtained from the camera were processed using the open-source pose estimation software programs OP version 1.7.0 and DLC version 2.3.2.

OP's "BODY_25" Pre-Trained network (OPPT) generated data for 25 keypoints (Fig 1B; see also Fig 9A in Supplementary Materials), whereas DLC employed the "Model Zoo full_human" Pre-Trained network (DLCPT) and generated data for 14 keypoints (Fig 1B; see also Fig. 9B 9A in Supplementary Materials).

DLC Custom-Trained model (DLCCT) was developed by custom training the artificial neural network "ResNet-101" pre-trained on the MPII Human pose dataset, a large-scale dataset of over 25,000 images containing over 40,000 individuals with labeled body parts (Andriluka et al., 2014a, Andriluka et al., 2014b, Insafutdinov et al., 2016). The "ResNet-101" is composed of deep convolutional and deconvolutional layers to predict the learned locations of selected keypoints employing feature detection (He et al., 2016). During the custom training process, the neural network learns to recognize and classify diverse patterns and objects, adjusting its weights to enhance accuracy in identifying features and patterns within the images (He et al., 2016). The dataset to train the neural network consisted of a total of 400 images and it was built by extracting 10 frames from each of the 40 videos using the k-means clustering. The k-means algorithm was used to ensure the dissimilarity of the chosen frames to optimize data clustering. Afterwards, a single operator (GP) manually applied, on each of the 400 images, 16 keypoints to the joint center locations of the shoulders, elbows, wrists, hips, knees, ankles, heels, and toes (Fig 1B; see also Fig 9C in Supplementary Materials). The 95% of the labeled images were used to train the DLC model, whereas the remaining 5% was used to assess the generalizability of the model (Mathis et al., 2018, Nath et al., 2019). The network was trained for 800,000 iterations using a personal computer (GPU: NVIDIA GeForce 3060, 12 GB). Also, to optimize the training of the neural network, the batch size was set to 1 and the data augmentation process was implemented to increase the diversity and quantity of training data, which is necessary to obtain a robust model (Mathis et al., 2018). This data augmentation process allows to scale the image contrast and apply the "Imgaug" package, which provides image random rotations ($\pm$ 25°) and image scaling (range: 0.5-1.25) (Wrench & Balch-Tomes, 2022). The training process lasted approximately 10 hours.

Once the training procedure ended, to quantify the error across learning, snapshots every 50,000 iterations were stored, and the Mean Average Euclidean error (MAE) between the manual labels and those predicted by DLCCT was evaluated. This technique not only enables the assessment of labeling accuracy in pixels but also prevents, via the early stopping method, the overfitting of the model by monitoring the MAE on the test, i.e., a subset comprising 5% of the data (Ratner, 2017, Mathis et al., 2018). Overfitting occurs when a model excels in performance on the training dataset yet struggles to generalize effectively to unseen data (test data). For this reason, only the snapshot with the lowest test MAE was chosen for additional analysis (Mathis et al., 2018, Nath et al., 2019).

In order to reach satisfactory performance for this locomotion task, the minimum number of images required was determined by repeating the training process with smaller training sets (300, 200 and 100 frames; DLCCT300, DLCCT200 and DLCCT100, respectively; Fig 1C; see Nath et al., 2019, Cronin et al., 2019).

Additionally, to further enhance the DLCCT performance, a refinement process was implemented as suggested by the DLC authors (Mathis et al., 2018; Nath et al., 2019). With this aim, from each dataset's training output two outlier frames per video (frames labeled by the model that contain errors) were manually corrected and then added to the initial training datasets (300 + 80 = 380, 200 + 80 = 280 and 100 + 80 = 180 frames; DLCCT380R, DLCCT280R and DLCCT180R, respectively; Fig 1C). After incorporating these corrections, the network underwent re-training for a total of 800,000 iterations (Nath et al., 2019). Also, the test MAE results of both DLCCT initial and refined datasets were used to compare the performances of the models.

The output files from the three markerless systems (OPPT, DLCPT, and DLCCT), which included x-y coordinates of the tracked body parts for each participant, were imported into an originally developed Python script for the further steps of analysis. Body parts coordinates were filtered using lowpass filter (zero-phase lag, fourth-order Butterworth) at 5 Hz (Winter, 2009). Data points with an accuracy estimated by the models of less than 0.5 were excluded, and then a linear interpolation was performed. To obtain gait temporal parameters, the HC and TO events were identified based on the displacement of x-coordinate of the heel and toe.

*Gait Parameters Estimations*

The HC and TO events obtained from both force platforms and pose estimation systems were used to compute the following temporal parameters (Stenum et al., 2021a):
- step time (the time interval between successive HCs of opposite feet);
- cadence (the number of steps per minute);
- stance time (the time interval between HCs and TOs of the same foot);
- double support time (the total amount of time spent with both feet in contact with the ground during a gait cycle).

*Statistical Analysis*

The mean, standard deviation (SD) and coefficient of variation (CV) of each gait parameter across pose estimations and force platforms were calculated. The Shapiro-Wilk test was adopted as a preliminary analysis to assess the Gaussian distribution of data (Razali & Wah, 2011).

The accuracy of the three markerless systems (OPPT, DLCPT and DLCCT) relative to the force platform were tested by using Pearson correlation coefficient, Bland-Altman analysis and absolute error. Pearson's correlation coefficient was computed to investigate the potential linear relationship between measurements estimated by the markerless system and the reference system. Bland-Altman analysis was employed to quantify the amount of agreement between systems of measurement by constructing 95% limits of agreement (LOA), which provides an estimate of the interval where 95% of the differences between both systems fall (Giavarina, 2015; Gerke, 2020). Absolute error was computed to assess the accuracy (i.e., the divergence of pose estimation output from reference system) and the precision (i.e., the consistency between the estimations) of the markerless systems. Absolute error was calculated, for each video *n,* by subtracting the reference system value from the pose estimation value.

$$Absolute\ error_n = |X_n estimated - X_n reference|$$

Where *X* represents the value of the temporal gait parameter.

Mean and standard deviation of the absolute error were calculated to quantify accuracy and precision, respectively.

$$Accuracy = \mu\ (Absolute\ error)$$

$$Precision = \sigma\ (Absolute\ error)$$

## Results

The DLCCT model performance in terms of test MAE are illustrated in Figure 2. The DLCCT100 exhibited the highest test MAE, followed by subsequent initial datasets (DLCCT200, DLCCT300, DLCCT400, respectively), showing that the test MAE reduced as the size of the dataset increased. Notably, when considering the refined datasets, each dataset had a test MAE markedly lower than its initial counterpart as well as the subsequent initial dataset. Specifically, DLCCT180R outperformed DLCCT100 and DLCCT200; DLCCT280R outperformed DLCCT200 and DLCCT300; and DLCCT380R outperformed DLCCT300 and DLCCT400.

The mean ± SD and CV% values of all temporal gait parameters for each system are reported in Table A of the Supplementary Materials.

The Shapiro-Wilk test, which was performed separately for each pose estimation and force platforms dataset, provided evidence supporting the assumption of a normal distribution within our data.

Pearson correlation analysis (Fig. 3A) showed a strong linear relationship (correlation coefficient greater than 0.7) with the reference system for all gait parameters across the following markerless systems: OPPT, DLCCT180R, DLCCT200, DLCCT280R, DLC300, DLC380R, and DLC400. The only parameters that exhibited a Pearson coefficient below 0.5 (i.e., indicating a low correlation) were the double support times of DLCPT and of DLCCT100, whereas the stance time of DLCPT had a correlation coefficient of approximately 0.5 (Fig 3A). The DLCCT380R demonstrated superior performance in comparison to the other DLCCT datasets (Fig 3B). Then, the correlation values obtained from the off-the-shelf systems (i.e., OPPT and DLCPT) were compared with those achieved with the best custom-trained DLCCT dataset for each gait parameter. As a result, DLCCT380R emerged again as the system with the stronger linear relationship with the reference system.

Bland-Altman plot analysis between force platforms and pose estimations systems measurements are presented for the step time in Figure 4, for the cadence in Figure 5, for the stance time in Figure 6, and for the double support time in Figure 7 (see also Table B in Supplementary Materials).

We can observe at first that the DLCCT LoAs tended to narrow as the initial dataset size increased, reaching the smallest dispersion with DLCCT400 for all the gait parameters. However, when considering the refined datasets, they consistently outperformed their corresponding initial datasets (i.e., DLCCT380R outperformed DLCCT300, DLCCT280R surpassed DLCCT200, and DLCCT180R surpassed DLCCT100). In addition, the refined datasets were also more efficient than their subsequent datasets (i.e., DLCCT380R outperformed DLCCT400, DLCCT280R outperformed DLCCT300, and DLCCT180R outperformed DLCCT200).

When comparing off-the-shelf systems, OPPT consistently demonstrated higher efficiency than DLCPT and outperformed the DLCCT sets up to DLCCT200. Indeed, DLCCT datasets beyond DLCCT200 became more performant (Figs. 4 – 7), with only two small exceptions found: the DLCCT180R cadence (Fig. 5) and DLCCT300 double support time (Fig. 7; see also Table B in Supplementary Materials).

Notably, DLCPT and DLCCT100 values display the widest LoAs for all gait parameters (Figs 4 – 7). LoAs of DLCCT100 exceed those of DLCPT in step time and cadence (Figs. 4 and 5); conversely in stance time and double support time DLCPT LoAs exceed those of DLCCT100 (Figs. 6 and 7).

The distributions of absolute errors for each markerless system with respect to the reference system are provided in Figure 8 (see also Table C in Supplementary Materials).

Across all the systems, DLCCT280R, DLCCT300, DLCCT380R, and DLCCT400 exhibited superior accuracy (white dots in Fig. 8) and precision (yellow vertical segments in Fig. 8), with DLCCT380R outperforming all the others. OPPT markedly outperformed DLCPT and DLCCT100 in all the considered time parameters. Also, OPPT displayed accuracies between DLCCT180R and DLCCT200, and precisions between DLCCT180R and DLCCT300 in step time, cadence, and stance time. In double support time OPPT achieved a better accuracy (between DLCCT280R and DLCCT300) and precision (between DLCCT180R and DLCCT300, see also Table C in Supplementary Materials). DLCPT and DLCCT100 exhibited different degrees of performance: DLCPT demonstrated higher accuracy in step time and cadence than DLCCT100, while DLCCT100 demonstrated a superior accuracy in stance time and cadence compared to DLCPT.

# Discussion

In this study we tested OP and both pre- and custom- trained models of DLC to measure temporal gait parameters. To the best of our knowledge, this is the first study to evaluate and compare the performance of both pre-trained and custom-trained DLC models during locomotion, while also providing insights into the refinement function of DLC.

Our analyses aimed to assess the performance of DLCCT models trained with both initial and refined datasets (through test MAE). Additionally, we evaluated the accuracy in terms of linear relationship (through Pearson correlation analysis), degree of agreement (through Bland-Altman analysis) and magnitude of the differences (through absolute error analysis) between all pose estimation and the reference systems.

The DLCCT test MAE showed a consistent reduction as the initial dataset frames increases, underscoring the importance of dataset size. Moreover, the implementation of the refinement function significantly contributes to minimizing test MAE and enhancing the precision of the DLCCT model, leading to a general optimization of the system that outperformed models trained with initial datasets.

Pearson correlation analysis of gait parameters (see Fig. 3A) reveals ranking results, where in the first group the DLCCT380R showed the strongest correlation with the reference system, followed by the DLCCT400, DLCCT280R, and DLCCT300. OPPT occupies an intermediate position, along with DLCCT180R and DLCCT200. Finally, DLCPT and the DLCCT100 exhibit the weakest correlations. These findings suggest that the accuracy and reliability of temporal gait parameters measurements are thus influenced not only by the choice of the pose estimation model but also prone to factors such as the dataset size and the use of refinement function.

The Bland-Altman analysis highlighted a narrowing of the LoAs as the initial DLC dataset size increased, further emphasizing the pivotal role of dataset size in influencing system performance (see Figs. $4 - 7$). The LoAs widths of the refined datasets were smaller than those of their corresponding initial and subsequent datasets, also supporting the Pearson correlation results (see Figs. $4 - 7$). In specific parameters such as step time and cadence, DLCPT surpasses DLC100, whereas in stance time and double support time DLC100 outperforms DLCPT. The poor performance of DLCPT in the stance time and double support time, which require identification of toe and heel positions, can be attributed to its sole detection of ankle movements compared to the other systems capturing both heel and toe keypoints, significantly affecting accuracy (see Fig. 9B in Supplementary Materials). These outcomes underscore how the impact of selecting appropriate keypoints influences the accuracy of gait parameter measurements.

The absolute error analysis, consistent with the trends observed in Pearson correlation and Bland-Altman analysis, confirmed the superior performance of DLCCT380R compared to all other systems (see Fig. 8). Moreover, this analysis further emphasizes the aforesaid performance disparities in performance between DLCPT and DLC100.

Our findings are consistent with recent studies (Needham et al., 2021, Washabaugh et al., 2022), wherein DLCPT and OPPT were compared with a marker-based motion capture system, that showed the inferior performance of DLCPT compared to that of OPPT. However, these studies did not consider the custom-trained model of DLC. Our results demonstrate that DLCCT outperforms both its pre-trained counterpart (with at least 200 labeled frames) and OPPT (with at least 300 labeled frames) in measuring temporal gait parameter, which underscore the importance of customizing the network for precise pose estimation.

Previous studies dealing with different motor behavior paradigms had already highlighted the potentials of the custom-trained DLC model (Moro et al., 2020, Vonstad et al., 2020, Drazan et al., 2021, Lonini et al., 2022, Haberfehlner et al., 2023). Considering this background, it is important to conduct a thorough analysis and comparison of DLC custom-trained and pre-trained models in order to validate and quantify the performance of this approach in various application fields.

It should be considered, however, that building a custom-trained model is a time-consuming process. Nevertheless, this strategy becomes necessary in situations where the pre-trained models are not suitable due to population characteristics (such as, for example, individuals in wheelchairs), specific motor tasks, or setup discrepancies not included in the pretrained model. In these cases, investing time and effort into customizing the model becomes essential to ensure accuracy and reliability of data.

Finally, our results demonstrate that the refinement function (Nath et al., 2019) provided by the DLC software reduces the need for manual labeling of new frames, decreases time consumption (Weber et al., 2022), and optimizes performance. Indeed, we conducted a novel comparison by matching the outcomes of the refinement function with those resulting from training with no refinement implementation. The comparison markedly favored the refinement function, revealing a significant improvement in accuracy compared to the initial datasets.

This study is not exempt from limitations. One limitation is the potential impact of a limited frame rate on the accuracy of our measurements (Fallahtafti et al., 2021). However, our goal was to implement and examine the performance of a low-cost clinical setting (Stenum et al., 2021, Viswakumar et al., 2022). For this reason, video recordings were temporally synchronized with platforms recordings to detect and validate the setup accuracy against an acknowledged reference system (Needham et al., 2021).

Our analysis for systems accuracies focused on temporal gait parameters to overcome the lack spatial parameters outputs associated with measurements conducted using traditional force platforms, which solely measure the vertical forces. Temporal gait parameters, derived from heel and toe events, are anyway fundamental for the study of human movement in clinical contexts and are habitually sufficient for effective evaluations and monitoring (Alaqtash et al., 2011, Mohammed et al., 2016, Tunca et al., 2017). For instance, the percentages of stance and swing time have been used to assess the subject's locomotion symmetry (Lopez-Meyer et al., 2011). Percentages of stance and swing time also serve as quantifiable indicators for measures of motor recovery after stroke and offer valuable feedback to both the patient and therapist (Lopez-Meyer et al., 2011, Chisholm et al., 2014).

## Conclusion

Our findings, in line with previous studies, confirm that OPPT proves to be the superior off-the-shelf markerless system, surpassing the DLCPT model. However, the DLCCT model outperforms both OPPT and DLCPT in the context of locomotion in healthy subjects, emphasizing the key role of customizing the networks for accurate pose estimations.

Furthermore, our results underscore the significant impact of the refinement function in the optimization of DLCCT performance allowing for higher accuracy with less time consumption. In summary, our findings contribute to the progress of knowledge and practical application of up-to-date open-source pose estimation systems for human movement analysis.

## Acknowledgements

We would like to thank Alessio Martinelli for his assistance in organizing the experimental sessions. Also, we extend our gratitude to Eros Quarta for his valuable suggestions.

## Author contributions

Study conception and design by G.P., R.B. Data acquisition by G.P., S.Gu. Data Analysis by G.P., S.Gr. Data interpretation by G.P, S.Gr. Figure design by G.P, D.M. Manuscript preparation G.P. All authors (G.P., S.Gr., S.Gu., L.M., D.M., R.B.) reviewed the manuscript.

# FIGURES

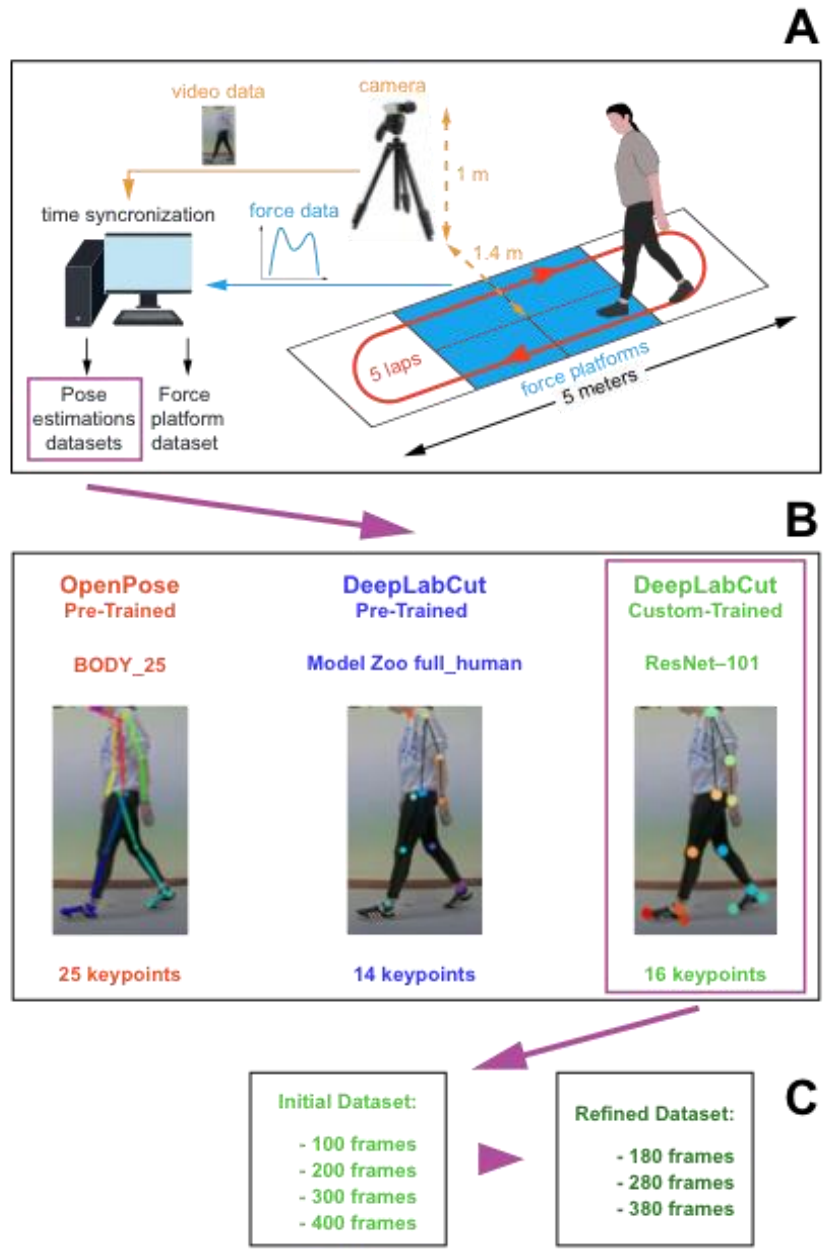

**Figure 1. (A)** Experimental setup and design. **(B)** OpenPose Pre-Trained, DeepLabCut Pre-Trained and DeepLabCut Trained software with their corresponding models (BODY_25, Model Zoo full_human, and ResNet-101, respectively) and keypoints illustration. **(C)** Initial trained datasets and refined datasets of DeepLabCut Trained. The color code used in B and C is maintained for all figures to represent the different systems.

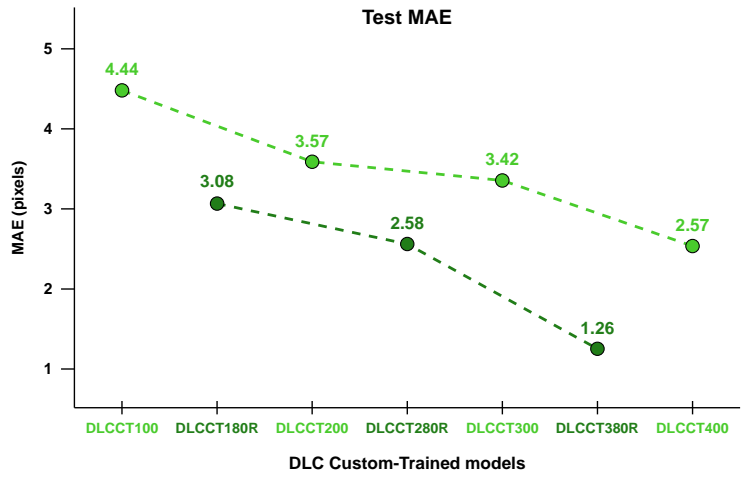

**Figure 2.** Test Mean Average Euclidean Error (test MAE) provided by DLC software for initial and refined datasets. Dots, data and connecting dashed lines are color coded.

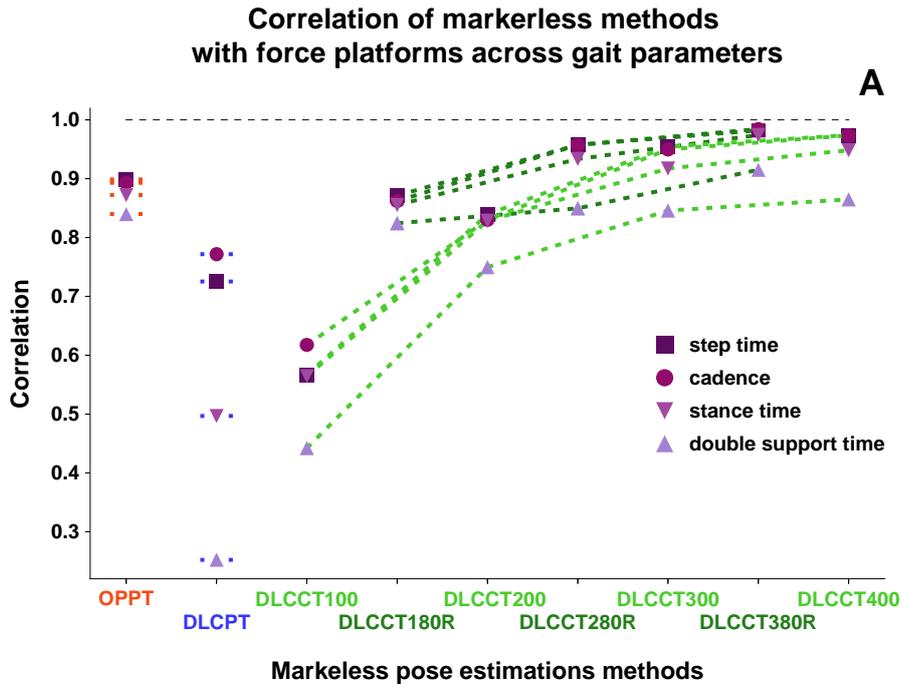

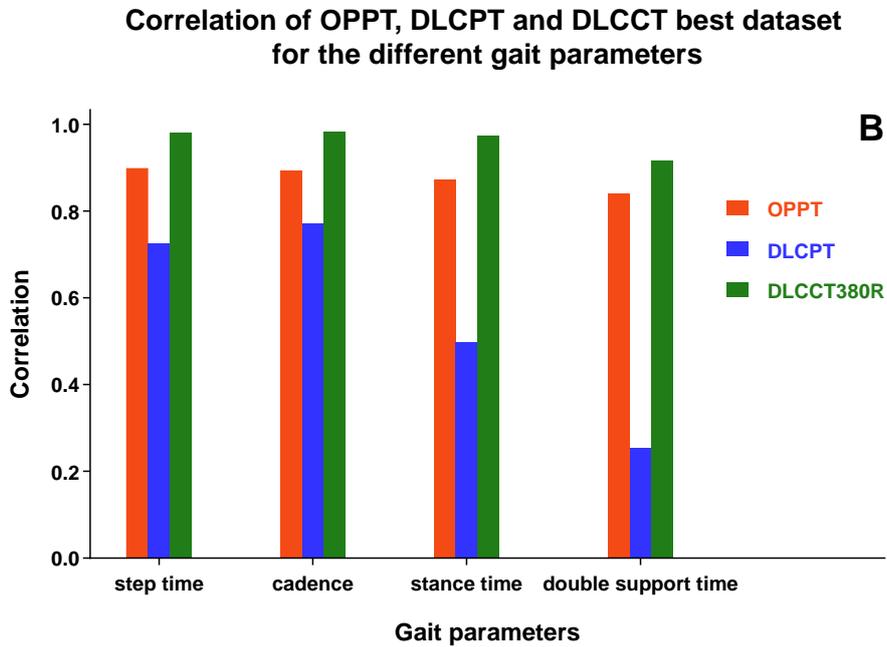

**Figure 3.** (**A**) Pearson correlation analysis of each markerless pose estimation system with force platforms (as the reference system), across gait parameters. (**B**) Pearson correlations of gait parameters for the most performant DLCCT dataset (i.e., DLCCT380R) as compared to the homologous OPPT, DLCPT values.

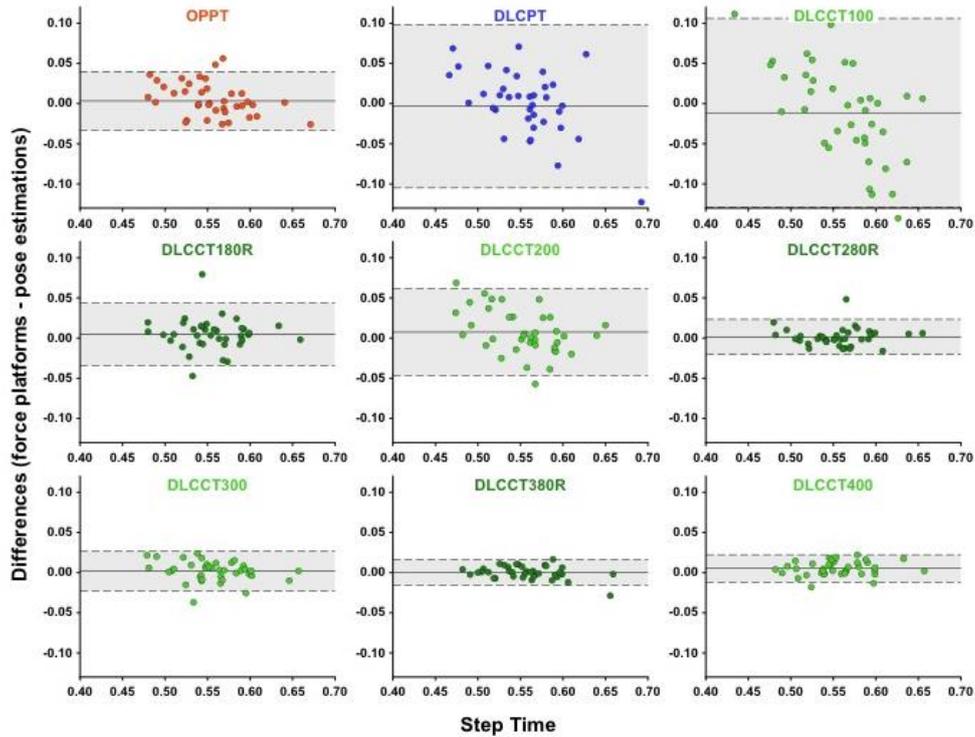

**Figure 4.** Bland-Altman analysis of OPPT, DLCPT, and of the different sets of DLCCT are illustrated for the step time. Dotted lines and enclosed gray areas indicate 95% LoA while the solid central line indicates the mean differences between force platforms and the pose estimation data per individual subject (color coded dots).

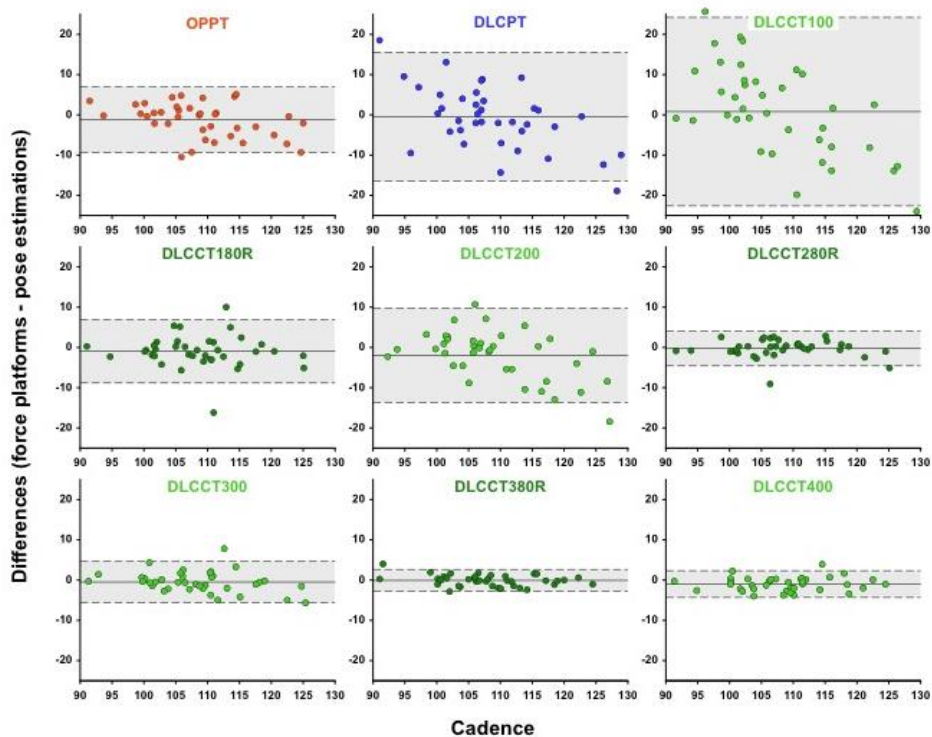

**Figure 5.** Bland-Altman analysis of OPPT, DLCPT, and of the different sets of DLCCT are illustrated for the cadence. Dotted lines and enclosed gray areas indicate 95% LoA, while the solid central line indicates the mean differences between force platforms and the pose estimation data per individual subject (color coded dots).

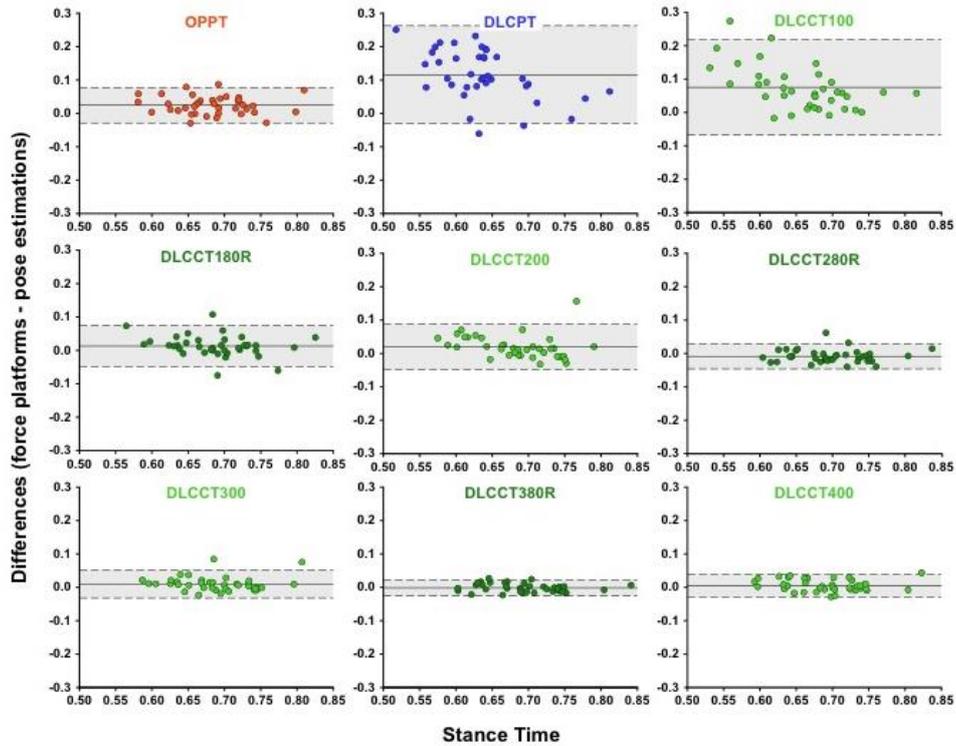

**Figure 6.** Bland-Altman analysis of OPPT, DLCPT, and of the different sets of DLCCT are illustrated for the stance time. Dotted lines and enclosed gray areas indicate 95% LoA, while the solid central line indicates the mean differences between force platforms and the pose estimation data per individual subject (color coded dots).

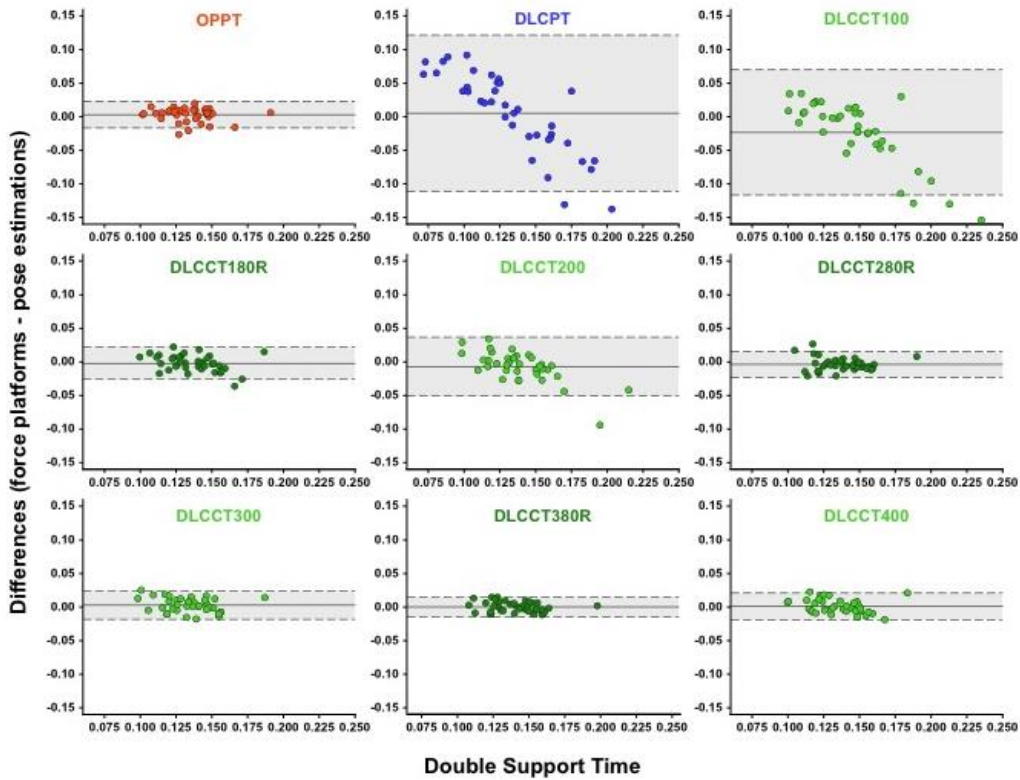

**Figure 7.** Bland-Altman analysis of OPPT, DLCPT, and of the different sets of DLCCT are illustrated for the double support time. Dotted lines and enclosed gray areas indicate 95% LoA, solid central line indicates the mean differences between force platforms and the pose estimation data per individual subject (color coded dots).

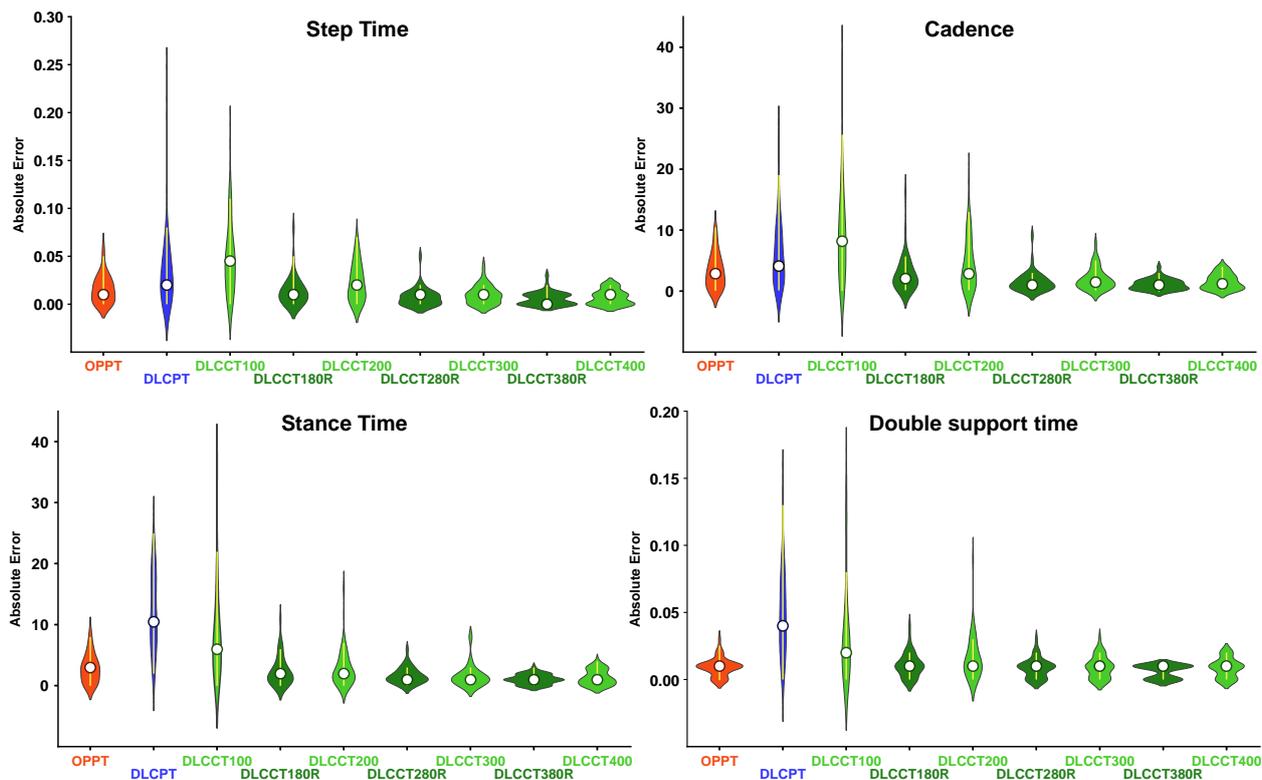

**Figure 8.** Violin plots display the distribution of absolute errors for each gait parameter across markerless systems. Within the single plots, white dots represent the accuracy (mean), while yellow vertical segments denote the precision (standard deviation).

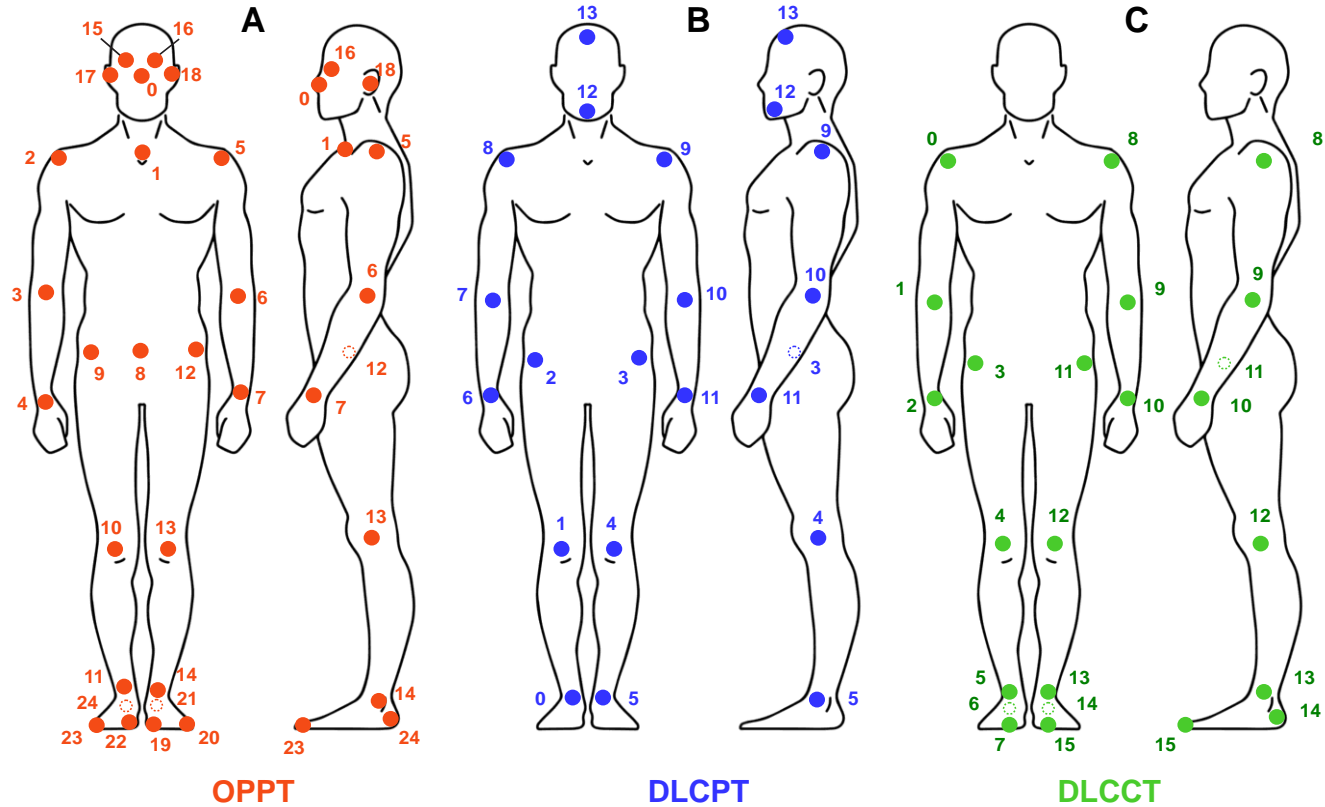

**Figure 9.** Keypoints visualization and corresponding anatomical landmarks for each markerless pose estimation system. (**A**) OPPT: 0 – Nose, 1 – Neck, 2 - Right Shoulder, 3 - Right Elbow, 4 - Right Wrist, 5 - Left Shoulder, 6 - Left Elbow, 7 - Left Wrist, 8 - Mid Hip, 9 - Right Hip, 10 - Right Knee, 11 - Right Ankle, 12 - Left Hip, 13 - Left Knee, 14 - Left Ankle, 15 - Right Eye, 16 - Left Eye, 17 - Right Ear, 18 - Left Ear, 19 - Left Big Toe, 20 - Left Small Toe, 21 - Left Heel, 22 - Right Big Toe, 23 - Right Small Toe, 24 - Right Heel. (**B**) DLCPT: 0 - Right Ankle, 1 - Right knee, 2 - Right Hip, 3 - Left hip, 4 - Left knee, 5 - Left ankle, 6 - Right wrist, 7 - Right elbow, 8 - Right shoulder, 9 - Left shoulder, 10 - Left elbow, 11 - Left wrist, 12 – chin, 13 – forehead. (**C**) DLCCT: 0 - Right Shoulder, 1 - Right Elbow, 2 - Right Wrist, 3 - Right Hip, 4 - Right Knee, 5 - Right Ankle, 6 - Right Heel, 7 - Right Toe, 8 - Left Shoulder, 9 - Left Elbow, 10 - Left Wrist, 11 - Left Hip, 12 - Left Knee, 13 - Left Ankle, 14 - Left Heel, 15 - Left Toe.

| System | Step time (s) | | | Cadence (steps/min) | | | Stance time (s) | | | Double support time (s) | | |
|---|---|---|---|---|---|---|---|---|---|---|---|---|
| | Mean | ± SD | CV% | Mean | ± SD | CV% | Mean | ± SD | CV% | Mean | ± SD | CV% |
| OPPT | 0.552 | 0.047 | 8.514 | 109.419 | 9.253 | 8.454 | 0.670 | 0.055 | 8.209 | 0.133 | 0.019 | 14.286 |
| DLCPT | 0.560 | 0.073 | 13.036 | 108.627 | 12.593 | 11.590 | 0.579 | 0.086 | 14.853 | 0.130 | 0.062 | 47.692 |
| DLCCT100 | 0.569 | 0.074 | 13.005 | 107.348 | 15.218 | 14.178 | 0.619 | 0.089 | 14.378 | 0.159 | 0.053 | 33.333 |
| DLCCT180R | 0.553 | 0.040 | 7.233 | 109.118 | 7.902 | 7.240 | 0.682 | 0.061 | 8.944 | 0.138 | 0.022 | 15.942 |
| DLCCT200 | 0.550 | 0.051 | 9.273 | 110.042 | 10.628 | 9.660 | 0.675 | 0.063 | 9.333 | 0.143 | 0.033 | 23.077 |
| DLCCT280R | 0.556 | 0.039 | 7.014 | 108.428 | 7.709 | 7.111 | 0.703 | 0.054 | 7.681 | 0.140 | 0.018 | 12.857 |
| DLCCT300 | 0.556 | 0.042 | 7.554 | 108.610 | 8.323 | 7.660 | 0.684 | 0.054 | 7.895 | 0.133 | 0.020 | 15.038 |
| DLCCT380R | 0.557 | 0.042 | 7.540 | 108.225 | 7.938 | 7.336 | 0.696 | 0.056 | 8.046 | 0.136 | 0.019 | 13.971 |
| DLCCT400 | 0.552 | 0.038 | 6.884 | 109.128 | 7.416 | 6.799 | 0.689 | 0.056 | 8.128 | 0.135 | 0.020 | 14.815 |
| Platforms | 0.557 | 0.039 | 7.002 | 108.139 | 7.544 | 6.972 | 0.695 | 0.053 | 7.626 | 0.136 | 0.018 | 13.235 |

**Table A.** Mean, ± standard deviations (± SD), and percent coefficient of variation (CV%) of gait parameters across pose estimations systems and force platforms.

| System | Step time (s) | | | Cadence (steps/min) | | | Stance time (s) | | | Double support time (s) | | |
|---|---|---|---|---|---|---|---|---|---|---|---|---|
| | Mean difference (Bias) | 95% LoA | | Mean difference (Bias) | 95% LoA | | Mean difference (Bias) | 95% LoA | | Mean difference (Bias) | 95% LoA | |
| | | Lower limit | Upper limit | | Lower limit | Upper limit | | Lower limit | Upper limit | | Lower limit | Upper limit |
| OPPT | 0.005 | -0.035 | 0.046 | -1.280 | -9.448 | 6.887 | 0.024 | -0.029 | 0.078 | 0.003 | -0.017 | 0.023 |
| DLCPT | -0.003 | -0.104 | 0.099 | -0.488 | -16.538 | 15.562 | 0.116 | -0.031 | 0.262 | 0.006 | -0.111 | 0.122 |
| DLCCT100 | -0.011 | -0.129 | 0.106 | 0.792 | -22.633 | 24.216 | 0.075 | -0.067 | 0.218 | -0.023 | -0.117 | 0.070 |
| DLCCT180R | 0.007 | -0.046 | 0.060 | -1.488 | -12.139 | 9.163 | 0.031 | -0.111 | 0.174 | -0.002 | -0.026 | 0.022 |
| DLCCT200 | 0.007 | -0.046 | 0.061 | -1.902 | -13.628 | 9.824 | 0.019 | -0.049 | 0.087 | -0.007 | -0.051 | 0.037 |
| DLCCT280R | 0.001 | -0.021 | 0.023 | -0.289 | -4.593 | 4.016 | -0.010 | -0.051 | 0.031 | -0.004 | -0.023 | 0.015 |
| DLCCT300 | 0.001 | -0.024 | 0.025 | -0.234 | -5.202 | 4.735 | 0.011 | -0.028 | 0.049 | 0.002 | -0.019 | 0.023 |
| DLCCT380R | 0.000 | -0.016 | 0.016 | -0.086 | -2.845 | 2.673 | -0.002 | -0.026 | 0.023 | 0.0002 | -0.015 | 0.015 |
| DLCCT400 | 0.005 | -0.012 | 0.022 | -0.988 | -4.322 | 2.345 | 0.006 | -0.028 | 0.040 | 0.001 | -0.019 | 0.021 |

**Table B.** Bland Altman analysis of each pose estimation method in comparison to the reference system.

| System | Step time (s) | | Cadence (steps/min) | | Stance time (s) | | Double support (s) | |
|---|---|---|---|---|---|---|---|---|
| | Accuracy (μ) | Precision (σ) | Accuracy (μ) | Precision (σ) | Accuracy (μ) | Precision (σ) | Accuracy (μ) | Precision (σ) |
| OPPT | 0.016 | 0.015 | 3.329 | 2.852 | 0.027 | 0.024 | 0.010 | 0.009 |
| DLCPT | 0.033 | 0.039 | 6.151 | 5.501 | 0.123 | 0.063 | 0.050 | 0.033 |
| DLCCT100 | 0.048 | 0.038 | 9.249 | 7.713 | 0.078 | 0.072 | 0.035 | 0.039 |
| DLCCT180R | 0.014 | 0.016 | 2.839 | 3.048 | 0.024 | 0.024 | 0.011 | 0.009 |
| DLCCT200 | 0.023 | 0.020 | 4.545 | 4.389 | 0.028 | 0.029 | 0.016 | 0.017 |
| DLCCT280R | 0.008 | 0.010 | 1.596 | 1.656 | 0.016 | 0.013 | 0.009 | 0.008 |
| DLCCT300 | 0.009 | 0.011 | 1.978 | 1.766 | 0.017 | 0.018 | 0.010 | 0.009 |
| DLCCT380R | 0.006 | 0.007 | 1.095 | 0.902 | 0.011 | 0.008 | 0.006 | 0.005 |
| DLCCT400 | 0.007 | 0.008 | 1.536 | 1.244 | 0.013 | 0.012 | 0.009 | 0.007 |

**Table C.** Absolute errors in terms of accuracy and precision of each pose estimation method in comparison to the reference system.